\documentclass[reprint, cha]{revtex4-1}

 \usepackage{graphicx}

\usepackage{gensymb}
\usepackage{amssymb}
\usepackage{amsmath}
\usepackage{textcomp}

\usepackage{xspace}

\usepackage{setspace}

\usepackage{subfigure}
\usepackage{color}
 \usepackage{amsthm}

\begin{document}

\title{Gabor wavelets combined with volumetric fractal dimension applied to texture analysis}

\author{\'Alvaro Gomez Z.}
 	     \email{agomez423@gmail.com}
\affiliation{Scientific Computing Group, S\~ao Carlos Institute of Physics, University of S\~{a}o Paulo (USP),  cx 369 13560-970 S\~{a}o Carlos, S\~{a}o Paulo, Brazil - www.scg.ifsc.usp.br} 

\author{Jo\~ao B. Florindo}
 	     \email{florindo@ursa.ifsc.usp.br}
\affiliation{Scientific Computing Group, S\~ao Carlos Institute of Physics, University of S\~{a}o Paulo (USP),  cx 369 13560-970 S\~{a}o Carlos, S\~{a}o Paulo, Brazil - www.scg.ifsc.usp.br} 

\author{Odemir M. Bruno}
              \email{bruno@ifsc.usp.br}
\affiliation{Scientific Computing Group, S\~ao Carlos Institute of Physics, University of S\~{a}o Paulo (USP),  cx 369 13560-970 S\~{a}o Carlos, S\~{a}o Paulo, Brazil - www.scg.ifsc.usp.br}

\date{\today}

\begin{abstract}
Texture analysis and classification remain as one of the biggest challenges for the field of computer vision and pattern recognition. On this matter, Gabor wavelets has proven to be a useful technique to characterize distinctive texture patterns. However, most of the approaches used to extract descriptors of the Gabor magnitude space usually fail in representing adequately the richness of detail present into a unique feature vector. In this paper, we propose a new method to enhance the Gabor wavelets process extracting a fractal signature of the magnitude spaces. Each signature is reduced using a canonical analysis function and concatenated to form the final feature vector. Experiments were conducted on several texture image databases to prove the power and effectiveness of the proposed method. Results obtained shown that this method outperforms other early proposed method, creating a more reliable technique for texture feature extraction.
\end{abstract}


\maketitle

\section{Introduction}

Texture analysis and classification have a huge variety of applications. Although it has been widely studied it remains open for research and in fact, is one of the biggest challenges for the field of computer vision and pattern recognition. 
There are a lot of different methods to deal with texture analysis, which can be grouped into four classes: (i) structural methods - where textures are described as a set of primitives; (ii) statistical methods - textures are characterized by non-deterministic measures
of distribution, using statistical approach; (iii)model-based - textures are described as mathematical and physical modeling; and (iv) spectral methods, based on the analysis in the frequency domain methods, such as Fourier, cosine transform or wavelets.
In the last approach, lay one of the well known and very succeed texture method: the Gabor filter, in which a feature extraction enhancement is proposed in this work.

The Gabor filter was proposed by Dennis Gabor in 1946 and extended by 2D and applied to image textures by Daugman \cite{Daugman1985,Daugman1980} in the 80's. Daugman's work main motivation was to model mathematically the receptive fields (response of neuronal cells set) of the cortical cells in the primate brain. Besides the biological motivation, the Gabor Filter has a very good performance for texture processing and still remains one of the best methods for texture analysis. 
Gabor texture technique consists on the convolution of an image with several multi-scale and multi-orientation filters. For each convolution, a transformed space is created, and the feature extraction is performed in each space. Usually, the feature vector is composed concatenating the energy measure of each convoluted image \cite{Rajadell}. This way, each convoluted image is represented by a single statistical value that is far from representing adequately the rich information present in the Gabor space. This issue has motivated the research in the field and the proposal of this work.

One of the simplest Gabor enhancement was proposed by \cite{Bandzi,Clausi,Shahabi}, which uses other basic statistical descriptors that proves to work better than energy in some situations. Another approach proposed is the use of GLCM \cite{Haralick} applied over the convoluted images to extract simple features achieving good results. Tou et al\cite{Tou1,Tou2}, proposed a simple yet powerful method to calculate the covariance matrix of all the convoluted images. More recently, the success of the LBP operator \cite{Ojala} in several computer vision fields motivated the adaptation of this operator on the Gabor process yielding the best results found on the literature.

In addition fractal dimension has been successfully used in texture feature extraction \cite{BackesB12,BackesB09a}. The fractal descriptors represent the spatial relations between pixel intensities, even small changes between texture patterns produce significant changes on the signature. In this paper, we propose the use of volumetric fractal dimension to extract the fractal descriptors of the Gabor convoluted images with the use of canonical analysis to decorrelate the signature descriptors and reduce dimensionality. The introduced approach is validated using several image texture datasets, and the results analyzed and compared against the best feature extraction methods for Gabor space found in the literature.

The paper is split into 9 sections. Next section gives a short overview of the Gabor wavelets method. Section 3 presents a brief description of the different methods implemented to compare their performance against the proposed technique. Section 4 explains the Volumetric fractal dimension method in detail. Section 5 presents the combinational approach of Gabor wavelets with volumetric fractal dimension. Section 6,7 and 8 shows the experiments conducted and the results obtained. Finally, section 9 draws conclusions and future directions.

\section{Gabor Wavelets}

Since the discovery and description of the visual cortex cells of mammalian our understanding of how the human brain process texture has advanced enormously. Daugman \cite{Daugman1980,Daugman1985} shown that simple cells in the visual cortex can be modeled mathematically using Gabor functions. These functions \cite{Gabor46} approximate cortex cells using a fixed gaussian. Later, Daugman proposed a two-dimensional Gabor wavelet \cite{Daugman2} for its application on image processing and it has been widely used in the field for its biological and mathematical properties. The 2D Gabor function is a local bandpass filter that achieves optimal localization in both spatial and frequency domain and allows multi-resolution analysis by generating multiple kernels from a single core function.

The Gabor wavelets are generated by dilating and rotating a single kernel with a set of parameters. Based on this concept, we use the Gabor filter function as the kernel to generate a filter dictionary. The two-dimensional Gabor transform is a complex sine wave with frequency $W$ modulated by a Gaussian function. Its form in space $ g(x, y) $ and frequency domains $ G(u, v) $, is given by Eqs.\ref{Eq1} and \ref{Eq2}:

\begin{equation}
g\left(x,y\right)=\left(\frac{1}{2\pi {\sigma }_x{\sigma }_y}\right){\exp  \left[-\frac{1}{2}\left(\frac{x^2}{{\sigma }^2_x}+\frac{y^2}{{\sigma }^2_y}\right)+2\pi jWx\right]\ }
\label{Eq1}
\end{equation}

\begin{equation}
G\left(u,v\right)={\exp  \left\{-\frac{1}{2}\left[\frac{{\left(u-W\right)}^2}{{\sigma }^2_u}+\frac{v^2}{{\sigma }^2_v}\right]\right\}\ }
\label{Eq2}
\end{equation}

A self-similar filter dictionary can be obtained by dilating and rotating $g(x, y)$ using the generating function proposed in \cite{Manjunath}.

\begin{equation}
g_{mn}=a^{-m}g(x',y')
\end{equation}

Where $a>1$ and $m,n$ are integer values that specify the number of scales and orientations respectively $m=0,1,...,M-1$ and $n=0,1,...,N-1$, where $M$ represents the total number scales and $N$ the total number of orientations. The $x'$ and $y'$ parameters are defined by:

\begin{equation}
x'=a^{-m}\left(x{\cos  \theta \ }+y{\sin  \theta \ }\right)
\end{equation}

\begin{equation}
y'=a^{-m}\left(-x{\cos  \theta \ }+y{\sin  \theta \ }\right)
\end{equation}

Where $\theta = \frac{nk}{N}$, the scaling factor $a^{-m}$ is needed to ensure that the energy is independent from $m$. The parameters necessary to generate the dictionary could be selected empirically. However, in \cite{Manjunath}, the authors present a suitable method to compose a filter dictionary that ensures a maximum spectrum coverage with the lowest redundancy possible. Based on this approach, we use the following equations to describe how to obtain the ideal sigmas.

\begin{equation}
a={(\frac{U_h}{U_l})}^{\frac{1}{M-1}}
\end{equation}

\begin{equation}
\vartheta_{u}=\ \frac{(a-1)U_h}{(a+1)\sqrt{2{\ln  2\ }}}
\end{equation}

\begin{equation}
\vartheta_{v}=\frac{tan(\frac{\pi}{2N})[U_h - 2 ln(\frac{\vartheta_{u}^2}{U_h})]}{\sqrt{2 ln 2 - \frac{(2 ln 2)^2 \vartheta_{u}^2}{U_h^2}}}
\end{equation}

Where $W = U_{h}$ and $U_h$ and $U_l$ represent the minimum and maximum central frequencies respectively.

\section{Gabor descriptors}

\begin{figure*}[!htb]
\centering
	\includegraphics[width=0.8\textwidth, height=0.9\textwidth]{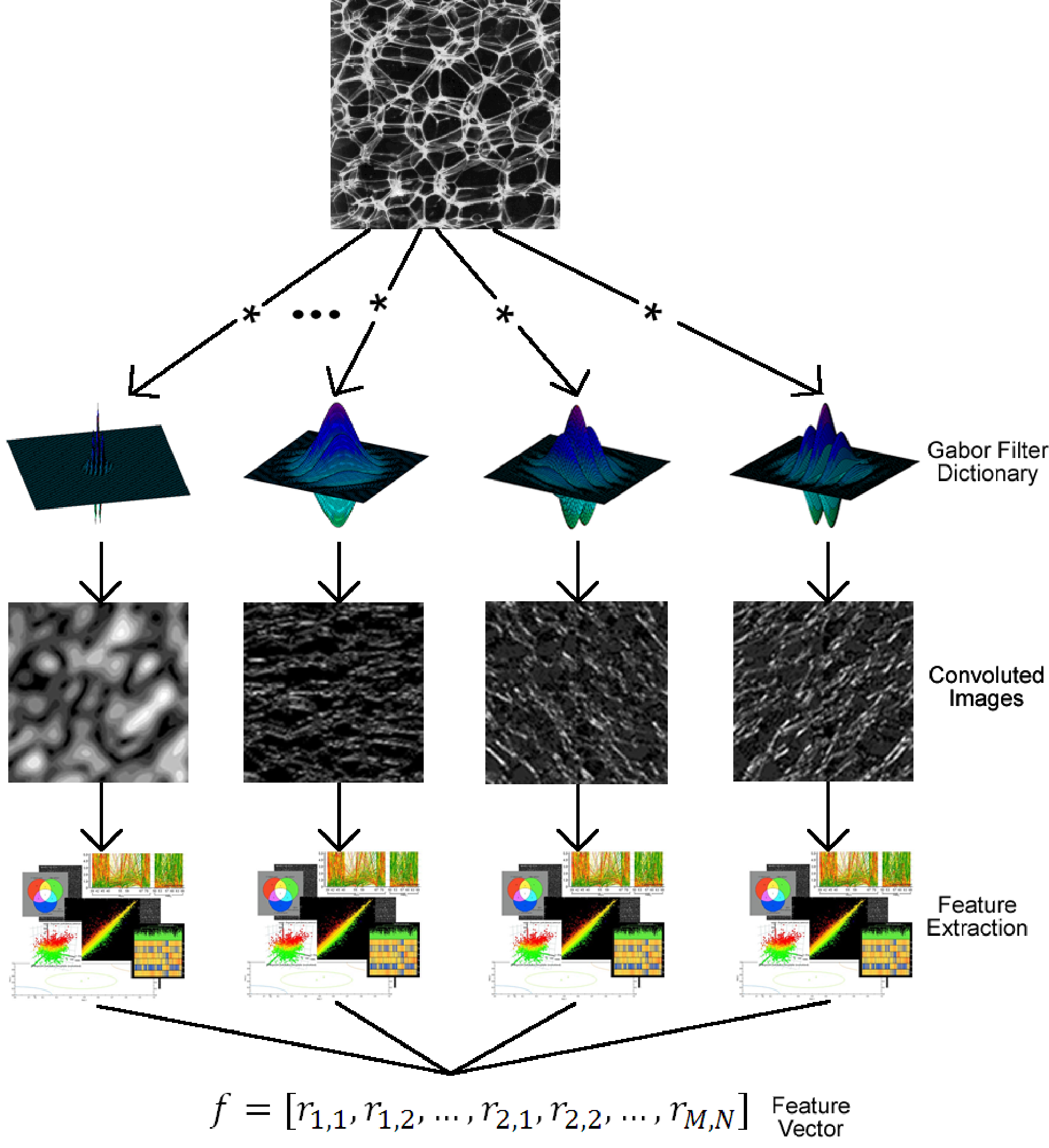}
\caption{General scheme used to extract features from the convoluted images.}
\label{fig:GaborProcess}
\end{figure*}

The Gabor wavelet representation of an image is the convolution of this image with the entire filter dictionary. Formally, the convolution result of an image $I(x,y)$ and a Gabor wavelet dictionary $\varphi_{f_u,m,n}$ named as \emph{Gabor images} on the rest of the paper can be defined as follows:

\begin{equation}
gi_{m,n}(x,y) = I(x,y) * \varphi_{f_u,m,n}(x,y)
\label{Eq:GaborConvolution}
\end{equation}

where $\varphi_{f_{u},m,n}$ denotes the Gabor wavelet with central frequency $f_{u}$, scale $m$ and orientation $n$. The number of images generated depends on the number of scales and orientations used. For example, four scales and six orientations will generate 24 Gabor images. The feature vector $F$ is composed by extracting single or multiple features from each generated image using image descriptors. A general process to describe this is shown in Figure \ref{fig:GaborProcess}.

A classical and simple approach to obtain the feature vector $F$ is just calculating the energy of each Gabor image by
\begin{equation}
F = [ e(gi_{1,1}), e(gi_{1,2}), ..., e(gi_{1,n}), e(gi_{2,1}), e(gi_{2,2}), ..., ..., e(gi_{m,n}) ]
\end{equation}
where $ e = \int \int f(x,y)^2 $ \cite{Daugman1985}. Although it is largely used in the literature, this approach does not achieve a efficiently information of the Gabor images. It has motivated the development of the methods to extract more efficiently the Gabor images information. In the following subsections a brief overview of the most important methods found on the literature is presented.

The non-orthogonal Gabor filters produce different effects depending on the texture characteristics. It does not exist an ideal combination of parameters that ensures the maximum performance. Whilst the work presented in \cite{Manjunath} help reducing the redundancy of the filters still some parameters like scales orientations and central frequencies are determined empirically. Thus, central frequencies variations seem to have a low impact on the results. They are fixed to $0.05$ and $0.4$ to reduce the number of variables for the experiments. In order to determine Gabor+method combination that obtains the best results for each combination, we performed eight experiments per method for each database. Each of those experiments represents a variation in the number of scales and orientations used in the Gabor wavelet process ranging from 2 to 6 scales and 3 to 6 orientations combined in an incremental framework: $2x6$, $3x4$, $3x5$, $4x4$, $4x6$, $5x5$, $6x3$, $6x6$ being scale x orientation.

For the purpose of comparison, the experiments are replicated using several state of the art techniques found in the related literature.

\subsection{Descriptors based on first order statistics}

Let $f(x,y)$ be a grayscale image with dimensions $x=0,1,...,W-1$ and $y=0,1,...,H-1$ where $W$ and $H$ are the image width and height respectively. The possible intensity values that $f(x,y)$ could take are $i=0,1,...,G-1$ where $G$ is the maximum number of intensity value. Then the histogram is a function showing the number of pixels for each possible grayscale intensity value according to:

\begin{equation}
h(i) = \sum_{x=0}^{W-1}\sum_{y=0}^{H-1}\delta(f(x,y),i)
\end{equation}

Where $\delta(i,j)$ its the binary function defined by:

\begin{equation}
\delta(j,i) = \left\{
\begin{array}{l l}
  1, & \quad \mbox{if j=i}\\
  0, & \quad \mbox{else}\\
\end{array} \right.
\end{equation}

Image histogram has the power to represent a large set of values in a single measure that reflects a specific property of the distribution. To compute descriptors, we use a histogram representation based on a density probability function given by:

\begin{equation}
p(i) = \frac{h(i)}{WH}, \quad i=0,1,...,G-1
\end{equation}

The density function $p(i)$ is a one-dimensional vector that holds important information that is later extracted using distribution measures such as energy, mean, variance, etc. The most common approach to extract features in the Gabor wavelets methods is energy based descriptors. Some recent approaches use other types of descriptors in order to obtain more useful information from each image. Since each extractor generates a single value from each image the final representation is a $(M$ x $N)$ - dimensional feature vector.

The best first-order statistics found in the literature are used on experimentation: Energy (Eq. 13), variance (Eq. 14) and percentil75 (Eq. 15) are used accordingly to their implementation in [34]. According to the figure \ref{fig:GaborProcess} the extractors are applied directly over the magnitude space.

\begin{equation}
E = \sum_{i=0}^{G-1}[p(i)]^2
\end{equation}

\begin{equation}
V = \sum_{i=0}^{G-1}(i-u)^{2}p(i)
\end{equation}

\begin{equation}
P_{75}=p_{ord} (\lceil 0.75(G-1) \rceil)
\end{equation}

Where $p_{ord}$ its the ascendant sorted vector of p and $u = \sum_{i=0}^{G-1}ip(i)$.

\subsection{Descriptors based on GLCM features}

Second-order statistics derived from the gray level co-ocurrence matrix (GLCM) are a better representation of how humans perceive texture patterns \cite{Clausi,Shahabi}. It has been proven to be the most successful approach to many kinds of texture feature extraction problems. GLCM features capture information regarding higher frequency components in texture. The co-ocurrence matrix represents the histogram of the number of occurrences of gray-level pair values when a pixel neighborhood algorithm is applied.

Formally, the GLCM $h_{d\theta} (i,j)$ represents the frequency of appearance of 2 pixels with gray-level values $a,b$ separated by a distance $d$ and orientation $\theta$ for an image $f(x,y)$ defined by:

\begin{equation}
f(x_1, y_1) = i \quad and \quad f(x_2, y_2) = j
\end{equation}

where

\begin{equation}
(x_2, y_2) = (x_1, y_1) + (d cos \theta, d sin \theta)
\end{equation}

For each $d$ and $\theta$ is created a squared matrix with a dimension the same size as the number of grayscale values present in the image, due to computational cost only a few values of $d$ and $\theta$ are used.

The research presented by \cite{Clausi} shows the finest combination of Gabor filters and gray level co-ocurrence matrix features. According to \cite{Clausi} these three basic statistic descriptors represent the best second order statistics extracted from the GLCM matrix obtained after processing the Gabor images:

\begin{equation}
Ent = - \sum_{i=0}^{G-1}\sum_{j=0}^{G-1}p(i,j)\log_2[p(i,j)]
\end{equation}

\begin{equation}
Con = \sum_{i=0}^{G-1}\sum_{j=0}^{G-1}(i-j)^2p(i,j)
\end{equation}

\begin{equation}
Cor = \sum_{i=0}^{G-1}\sum_{j=0}^{G-1}\frac{ijp_{d\theta}(i,j)-\micro_{x}\micro_{y}}{\sigma_x \sigma_y}
\end{equation}

\subsection{Descriptors based on covariance matrix features}

Covariance matrix is a statistical method that represents the covariance between values. Covariance matrix applied to images reflects important features of heterogeneous images while achieving a considerable dimensionality reduction. A covariance matrix can be represented as:

\begin{equation}
C_R = \frac{1}{n-1}\sum_{k=1}^{n}(z_k-u)(z_k-u)^T
\end{equation}

where z represents the feature point and $u$ the mean of $n$ feature points. For fast computation, integral image technique is used \cite{Tuzel}. The $P$ and $Q$ tensor used for the computation are defined by:

\begin{equation}
P(x',y',i) = \sum_{x<x',y<y'}F(x,y,i)   i=1...d
\end{equation}

\begin{equation}
Q(x',y',i,j) = \sum_{x<x',y<y'}F(x,y,i)F(x,y,i)   i,j=1...d
\end{equation}

where $F$, represents the feature image and $d$ the number of dimensions of the covariance matrix. Hence, 24 images generate a 24x24 matrix. Finally, the covariance matrix is generated using $P$ and $Q$.

\begin{equation}
\begin{split}
  & \left. C_{R(x',y';x'',y'')} = \frac{1}{N-1}[Q_{x'',y''}+Q_{x',y'}-Q_{x'',y'} \right. \\
  & \left. -\frac{1}{N}(P_{x'',y''}+P_{x',y'}-P_{x',y''}-P_{x'',y'})(P_{x'',y''}+P_{x',y'}-P_{x',y''}-P_{x'',y'})^T] \right.
\end{split}
\end{equation}

where $(x',y')$ is the upper left coordinate and $(x'',y'')$ is the lower right coordinate of the image.

The covariance matrix implementation follows the directives given in \cite{Tou2}. The final covariance matrix obtained has dimensions $K x K$ where $K= m x n$. Since the covariance matrix is a symmetric matrix, only the non repeated values from the matrix are used as features. Hence, a $24x24$ covariance matrix generates a feature vector of size $300$.

\subsection{Descriptors based on local binary pattern features}

Some of the latest work in Gabor signatures involves descriptors based on local binary patterns. The original LBP operator \cite{Ojala} labels the pixels of an image by thresholding the $3 x 3$ neighborhood of each pixel $f_p(p=0,1,2,...,7)$ with the center value $f_c$ and considering the result as a binary number according to:

\begin{equation}
S\left(f_p-f_c\right)=\ \left\{ \begin{array}{c}
1,\ \ \ \ \ f_p=f_c \\
0,\ \ \ \ \ f_p<f_c \end{array}
\right.
\end{equation}

Then, by assigning a binomial factor $2^p$ for each $S(f_p-f_c)$ the LBP pattern for each pixel is achieved as:

\begin{equation}
LBP=\sum^7_{p=0}{S\left(f_p-f_c\right)2^p}
\end{equation}

In \cite{Zhang} The LBP operator is applied to each pixel on the Gabor images to generate a LGBP map (Local Gabor Binary Map). $G_{lgbp}(x,y,u,v)$ the concatenation of the histograms of each Gabor image is used as the feature vector. In \cite{Shufu} a volume approach is taken by considering all the Gabor images as a 3D volume and performing a LBP calculation in the 3D space.

The local binary pattern is applied to the Gabor images according to \cite{Zhang}. A 4-neighbourhood is applied to reduce the size of the histogram. since a 4-neighbourhood allow a maximum of $16$ possible values on the LBP map $R$. The final feature vector is composed of the concatenation of the histogram of each Gabor image:

\begin{equation}
H = [h_{1,1},h_{1,2},...,h_{2,1},h_{2,2},...,h_{m,n}]
\end{equation}

where $m,n$ is the number of scales and orientations used for the Gabor process and $h$ is:

\begin{equation}
h_{1,1,i} = \sum_{x,y\in R}(I{G_{glbp}(x,y,u,v)=i)}
\end{equation}

\section {The proposed method}

\subsection{Volumetric fractal dimension}

The fractal concept was first used by Mandelbrot in his book \cite{Mandelbrot}. This concept states that natural objects cannot be described using Euclidean geometry but using persistent self-repeating patterns. In recent years this concept has been used on the field of image analysis \cite{BackesCB09a,BackesB12,BackesB09a,BrunoPFC08}. To adapt the fractal concept to images is necessary to use a measure that captures fractal properties of non fractal objects inside discrete environments. For this purpose, the fractal dimension of an image is used to describe how self-repetitive the objects contained within the image are. Under this concept, several types of images could be analyzed. An approach used to analyze grayscale images called volumetric fractal dimension proposed in \cite{BackesCB09a,BackesB09a,BackesJKB09} has proven to be a very effective fractal descriptor. On \cite{Gomez} the authors successfully demonstrated the power of VFD to describe the Gabor images. On this approximation, we take a different approach to reduce and de-correlate the fractal signatures in order to improve the power of description and reduce dimensionality.

Let $gi_{m,n}(x,y)$ be a Gabor image taken from Eq. \ref{Eq:GaborConvolution} the 3-dimensional representation necessary to compute the VFD is given by $S(x,y,z) \exists R^3$ where $(x,y)$ are the spatial coordinates of the image and $z$ is the gray level intensity. This surface $S$ is dilated by a sphere of radius $r$ and the influence volume of the dilated surface $V(r)$ is calculated for each value of $r$. This could be better explained by equation:

\begin{equation}
V(r) = \{p' \epsilon R^3 | \exists p \epsilon S:|p-p'| \leq r\}
\label{Eq:VFD}
\end{equation}

where $ p'=(x',y',z')$ is a point in $R^3$ whose distance from $p=(x,y,z)$ is smaller or equal to $r$. As $r$ grows the spheres start to intercept each other producing variations on the computed volume. This property makes VFD very sensitive to even small changes on the texture pattern. Each expansion of $r$ generates a single-volume measure. Therefore, the values that $r$ takes must reflect each possible state of the expansion without redundancy. To reduce the computational costs of the volume computation, we applied an exact 3-dimensional Euclidean distance transform algorithm (EDT) \cite{FabbriCTB08} over the surface. The EDT performs a calculus of the distance of all the voxels on $R^3$ to its closest $p' \exists R^3$ voxel using the Euclidean distance. The most suitable way to obtain the set of radius to expand the surface is by using all the possible Euclidean distances up to a maximum radius. This is defined by:

\begin{equation}
E={1,\sqrt{2},\sqrt{3},...,r_{max}}
\end{equation}

The fractal dimension can be estimated as

\begin{equation}
D= 3-\lim_{r->0}\frac{\log(V(r))}{\log(r)}
\end{equation}

The fractal signature (or fractal descriptors) will be composed by the logarithm of each volume according to:

\begin{equation}
F = [\log V(1),\log V(\sqrt{2}),\log V(\sqrt{3}),...,\log V(r_{max})]
\end{equation}

\begin{figure*}[!htb]
\centering
	\subfigure[]{\includegraphics[width=0.20\columnwidth, height=0.20\columnwidth]{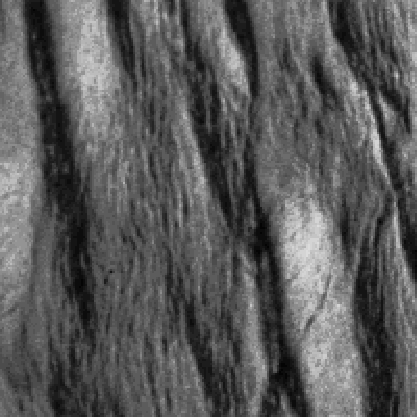}}
	\subfigure[]{\includegraphics[width=0.35\columnwidth, height=0.15\columnwidth]{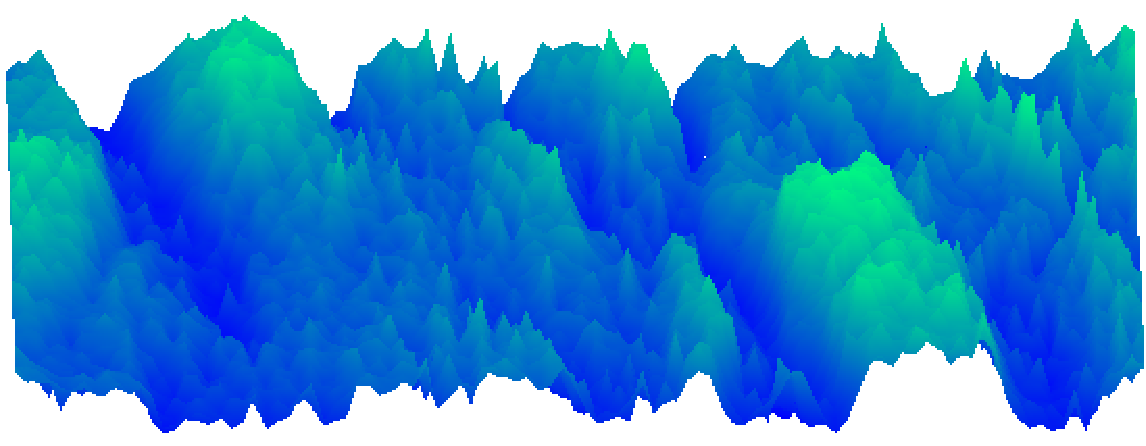}}
	\subfigure[]{\includegraphics[width=0.35\columnwidth, height=0.15\columnwidth]{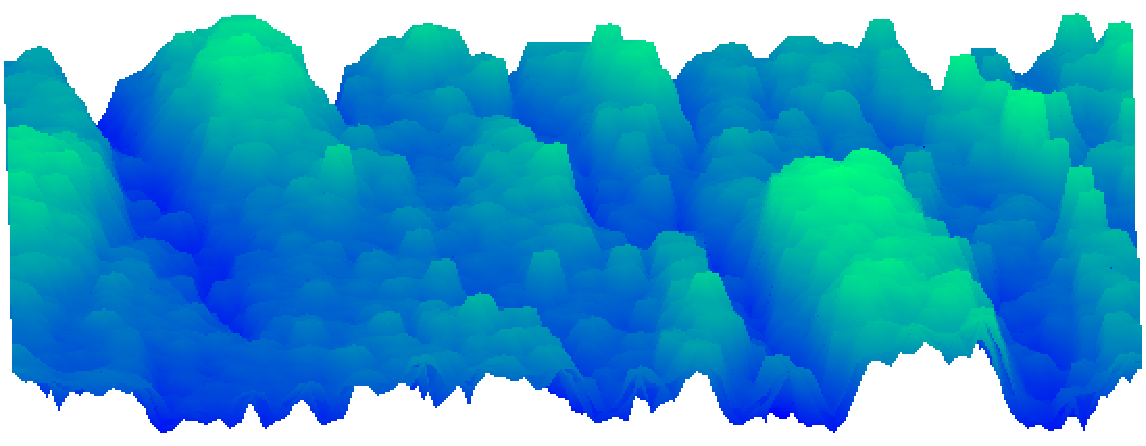}}
    \subfigure[]{\includegraphics[width=0.35\columnwidth, height=0.15\columnwidth]{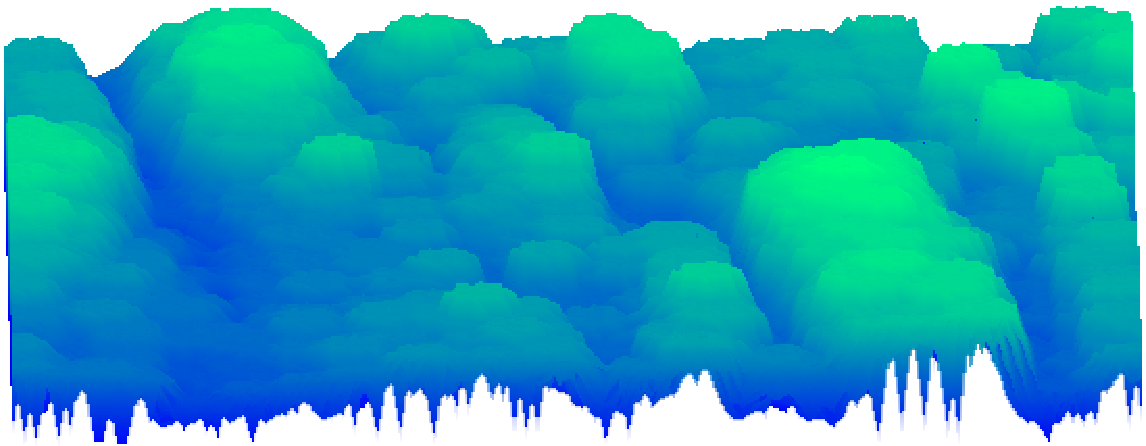}}
    \subfigure[]{\includegraphics[width=0.35\columnwidth, height=0.15\columnwidth]{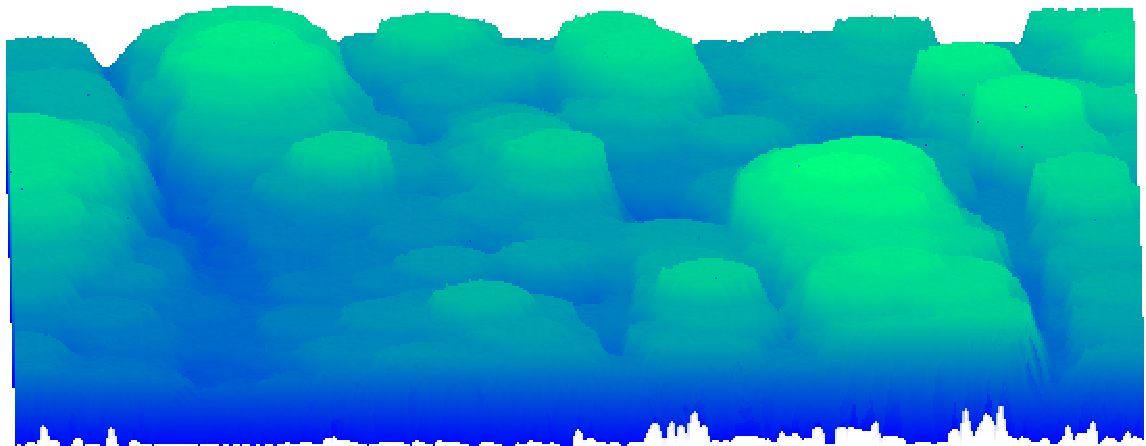}}
\caption{(a) is an image taken from Brodatz database, (b) image with expansion $r = 2$, (c) image with expansion $r = 5$ (d) image with expansion $r = 7$ (e) image with expansion $r = 9$}
\label{fig:Derivation}
\end{figure*}

The parameters used in the feature extraction are based on previous research presented by \cite{BackesB09a,BackesJKB09} where the expansion radius for the Volumetric Fractal dimension is set to 16. The number of canonical variables used is based on the percentage of representation of the i-th most important canonical variables. Volumetric fractal dimension signature tends to be $99.90\%$ described with only 10 canonical variables. Figure \ref{fig:Derivation} shows the process.

\subsection{Canonical discriminant analysis}

The Canonical discriminant analysis (CDA) is dimension reduction technique closely related to principal component analysis. CDA purpose is to find linear combinations of quantitative variables that provide maximal separation between classes \cite{McLachlan}. This linear combinations posses the power of producing a reduced number of independent features also called canonical variables.

The total dispersion among the feature vectors is defined as:

\begin{equation}
S = \sum^N_{i=1}(\varphi_i-\overrightarrow{M})(\varphi_i-\overrightarrow{M})'
\end{equation}

where $\overrightarrow{M}$ is the global mean feature vector and $\overrightarrow{\varphi_i}$ contains the row features of all vectors for class $i$, defined by:

\begin{equation}
\overrightarrow{\varphi_i} = F_M(x_\sigma)
\end{equation}

\begin{equation}
\overrightarrow{M} = \frac{\sum^N_{x=1}F_M(x)}{N}
\end{equation}

M is the total number of features. The matrix $S_i$ indicating the dispersion of objects within each class, is defined as:

\begin{equation}
S_i = \sum_{i \in C_i}(\overrightarrow{\varphi_i} - \overrightarrow{u_i})(\overrightarrow{\varphi_i} - \overrightarrow{u_i})'
\end{equation}

where $\overrightarrow{u_i}$ is the mean feature vector for objects in class $i$ defined by:

\begin{equation}
\overrightarrow{u_i} = \frac{\sum^N_{x=1}\overrightarrow{F_i}(x_k)}{N_k}
\end{equation}

The intraclass variability $S_{intra}$ indicates the combined dispersion in each class is defined by:

\begin{equation}
S_{intra} = \sum^K_{i=1}S_i
\end{equation}

The interclass variability $S_{inter}$ indicates the dispersion of the classes in terms of their centroids is defined by:

\begin{equation}
S_{inter} = \sum^K_{i=1}N_i(\overrightarrow{u_i} - \overrightarrow{M})(\overrightarrow{u_i} - \overrightarrow{M})'
\end{equation}

where K is the number of classes and N the number of samples on class $i$, Finally we have the total variability represented by:

\begin{equation}
S = S_{intra} + S_{inter}
\end{equation}

Finally, to obtain the principal components we use the approximation taken in \cite{RossattoCKB11}:

\begin{equation}
C = S_{inter}*S_{intra}^{-1}
\end{equation}

The i-th canonical discriminant function is given by:

\begin{equation}
Z_i = a_{i1}X_1 +  a_{i2}X_2 + ,,,+ a_{ip}X_p
\end{equation}

where p is the number of features and $a_{ij}$ are the sorted eigenvectors of C where $a_1$ is the most significant eigenvector. This definition leads to $Z_i$ non correlated features, where $i$ is the number of features used to reduce de dimensionality of the dataset with $i<p$.

\subsection{Proposed Signature}

Let $gi_{m,n}(x.y)$ be the convoluted image from equation \ref{Eq:GaborConvolution}. Let $e$ be the set of Euclidean distances $e=[1,\sqrt{2}, \sqrt{3},...,r_{max}]$ for a radius $r_{max}$. The volumetric fractal dimension signatures of each Gabor image $gi_{m,n}(x,y)$ is defined by:

\begin{equation}
\omega_{m,n}(z) = \{ VFD(gi_{m,n}(x,y),r)  | \forall r \in e \}
\end{equation}

where $r$ is a radius from vector $e$ and $\omega_{m,n}(z)$ is a vector that contains the fractal signatures for all the Gabor $m,n$ images. Then a canonical analysis function is applied to de-correlate the signature descriptors and $N$ principal components are selected. The computation of the canonical analysis of the signatures is defined by:

\begin{equation}
\phi_{m,n}(z) = \{\lambda(\omega_{m,n},N)\}
\end{equation}

Where $\{\lambda(\omega_{m,n},N)\}$ is the $N$ principal components of $\omega_{m,n}$ with orientation $m$ and scale $n$. Finally the image feature vector $F$ consists on the concatenation of the principal components previously computed defined by:

\begin{equation}
F = [ \phi_{1,1}(1), \phi_{1,1}(2), ...,\phi_{1,1}(z), \phi_{1,2}(1), \phi_{1,2}(2), ...,\phi_{1,1}(z),...,\phi_{m,n}(z) ]
\end{equation}

\section{Evaluation strategy}

\textbf{Image Databases: } For experimentation purposes, we used five different image databases. All the related methods and the proposed method are tested with each database. The image databases are selected based on the recurrence which each database is used in connected literature to validate feature extraction methods. The selection contains databases with a different difficulty level in classification and reported results. The selected databases were:

\begin{itemize}

\item Brodatz: Obtained from \cite{Brodatz} it contains 111 textures in grayscale each with 640 x 640 pixels. To generate a database with the appropriate number of samples per class, we took 10 non-overlapping random windows of 200 x 200 pixels from each texture, hence, the used database contains 1110 images with 111 classes and 10 images per class.

\item KTH-TIPS2: Obtained from \cite{Fritz} the "2b" version was selected and it contains 11 grascale textures each with 108 samples of 200 x 200 pixels.

\item  Outex texture classification test suite 5: Obtained from \cite{Outex5} the selected $Outex_TC_00005$ contains 24 grayscale textures each with 368 samples of 32 x 32 pixels.

\item  Outex texture classification test suite 5: Obtained from \cite{Outex5} the selected $Outex_TC_00014$ contains 68 grayscale textures each with 368 samples of 128 x 128 pixels.

\item  Outex texture classification test suite 5: Obtained from \cite{Outex5} the selected $Outex_TC_00016$ contains 319 grayscale textures each with 368 samples of 128 x 128 pixels.

\end{itemize}


\textbf{Classification: }  With the extracted features are possible to perform a class separation based on the use of a statistical classifier. We have chosen the use of naive Bayes classifier \cite{Mitchell} which is a simple probabilistic classifier based on the Bayes theorem. This classifier uses an independent feature model where the presence or absence of a particular feature of a class is unrelated to the presence of absence of any other feature. In simple terms, it assumes the conditional independence among attributes. Despite its over-simplified assumptions, this classifier has worked very well with the real world datasets even when the attribute independence hypothesis is violated \cite{Kuncheva}, \cite{Domingos}.

Formally, the probability of an observation $ E=(x_1,x_2,...,x_n)$ being class c is:

\begin{equation}
p(c|E) = \frac{p(E|c)p(c)}{p(E)}
\end{equation}

where E is the defined as the class C = + if:

\begin{equation}
f_b(E)=\frac{p(C=+|E)}{p(C=-|E)} \geq 1
\end{equation}

Where $f_b(E)$ is called a Bayesian classifier. based on the attribute independency hypothesis we can write

\begin{equation}
p(E|c)=p(x_1,x_2,...,|x_n|c)= \prod_{i=1}^n p(x_i|c)
\end{equation}

The resulting naive Bayes classifier can be defined as:

\begin{equation}
f_{nb}(E) = \frac{p(C=+)}{p(C=-)} \prod_{i=1}^n \frac{p(x_i|C=+)}{p(x_i|C=-)}
\end{equation}

Even though the Naive Bayes classifier still does a good job with non-independent features is not appropriate to use highly correlated features. To solve this problem, we use the canonical discriminant analysis function over de dataset to remove correlations. The application of this method maximizes the separation between classes and reduces de dimensionality of the dataset.

\section{Experimental results}

The results obtained for each image database is presented in this section. Each table shows the rate of correct classifications. All the techniques implemented for the purpose of comparison are run against all the image databases.

\begin{table*}[!h]
\begin{center}\small
\caption{Results for Brodatz image database.}
\begin{tabular}{ccccccccc}
\hline
                                                                        \multicolumn{ 9}{c}{ \textbf{Scales x Orientations} } \\
\hline
    &       &       &       &       &       &       &      &       \\
   \multicolumn{ 1}{c}{\textbf{Gabor +}} & \multicolumn{ 1}{c}{\textbf{2 x 6}} & \multicolumn{ 1}{c}{\textbf{3 x 4}} & \multicolumn{ 1}{c}{\textbf{3 x 5}} & \multicolumn{ 1}{c}{\textbf{4 x 4}} & \multicolumn{ 1}{c}{\textbf{4 x 6}} & \multicolumn{ 1}{c}{\textbf{5 x 5}} & \multicolumn{ 1}{c}{\textbf{6 x  3}} & \multicolumn{ 1}{c}{\textbf{6 x 6}} \\
\hline

    \textbf{Energy} & 62.59 &      79.22 &      80.58 &      84.71 &      86.48 &      86.00 &      83.13 &      87.54 \\

  \textbf{Variance} &     69.65 &	85.25 &	86.39 &	85.56 &	87.40 &	86.30 &	82.84 &	87.16 \\

\textbf{Percentil75} &     64.71	& 82.86	& 83.53	& 85.59	& 87.45	& 87.68	& 84.68	& 87.77 \\

\textbf{LBP} &      92.52	& 92.96	& 92.75	& 92.14	& 90.21	& 89.05	& 90.66	& 81.68 \\

      \textbf{Covariance} &      70.38	& 85.55	& 85.39	& 89.41	& 89.86	& 89.01	& 86.52	& 88.14 \\

      \textbf{GLCM} &      79.90	& 88.90	& 88.54	& 88.37	& 84.55	& 86.00	& 84.91	& 79.51 \\

      \textbf{Enhanced Fractal} &      93.51 &      94.05 &      93.88 &      94.05 &      \textbf{95.59} &      94.68 &      93.87 &      94.14 \\

\hline
\end{tabular}
\end{center}

\label{tab:Table1}
\end{table*}

\begin{table*}[!h]
\begin{center}\small
\caption{Results for KTH-TIPS2b image database.}
\begin{tabular}{ccccccccc}
\hline
                                                                        \multicolumn{ 9}{c}{ \textbf{Scales x Orientations} } \\
\hline
    &       &       &       &       &       &       &      &       \\
   \multicolumn{ 1}{c}{\textbf{Gabor +}} & \multicolumn{ 1}{c}{\textbf{2 x 6}} & \multicolumn{ 1}{c}{\textbf{3 x 4}} & \multicolumn{ 1}{c}{\textbf{3 x 5}} & \multicolumn{ 1}{c}{\textbf{4 x 4}} & \multicolumn{ 1}{c}{\textbf{4 x 6}} & \multicolumn{ 1}{c}{\textbf{5 x 5}} & \multicolumn{ 1}{c}{\textbf{6 x  3}} & \multicolumn{ 1}{c}{\textbf{6 x 6}} \\
\hline

    \textbf{Energy} & 55.84 &	72.64 &	72.50 &	75.11 &	75.69  &	73.21 &	70.67 &	73.24 \\

  \textbf{Variance} &     58.25 &	74.05 &	74.09 &	73.01 &	73.32 &	71.73 &	69.22 &	72.59 \\

\textbf{Percentil75} &     55.45 &	74.51 &	74.97 &	76.92 &	77.43 &	76.66 &	74.38 &	77.38 \\

\textbf{LBP} &      86.17 &	84.92 &	84.86 &	84.15 &	82.47 &	78.80 &	78.80 &	71.14 \\

      \textbf{Covariance} &      51.73 &	83.01 &	81.76 &	76.61 &	75.79 &	74.47 &	72.23 &	74.70 \\

      \textbf{GLCM} &      64.13 &	75.70 &	70.98 &	71.63 &	68.15 &	65.26 &	66.66 &	58.31 \\

      \textbf{Enhanced Fractal} &      90.49 &      89.90 &      90.40 &      90.32 &      \textbf{91.58} &      90.07 &      89.39 &      88.38 \\

\hline
\end{tabular}
\end{center}

\label{tab:Table2}
\end{table*}

\begin{table*}[!h]
\begin{center}\small
\caption{Results for Outext test suite 5 database.}
\begin{tabular}{ccccccccc}
\hline
                                                                        \multicolumn{ 9}{c}{ \textbf{Scales x Orientations} } \\
\hline
    &       &       &       &       &       &       &      &       \\
   \multicolumn{ 1}{c}{\textbf{Gabor +}} & \multicolumn{ 1}{c}{\textbf{2 x 6}} & \multicolumn{ 1}{c}{\textbf{3 x 4}} & \multicolumn{ 1}{c}{\textbf{3 x 5}} & \multicolumn{ 1}{c}{\textbf{4 x 4}} & \multicolumn{ 1}{c}{\textbf{4 x 6}} & \multicolumn{ 1}{c}{\textbf{5 x 5}} & \multicolumn{ 1}{c}{\textbf{6 x  3}} & \multicolumn{ 1}{c}{\textbf{6 x 6}} \\
\hline

    \textbf{Energy} & 50.77 &	68.74 &	69.32 &	72.31 &	73.94 &	75.73 &	72.94 &	77.08 \\

  \textbf{Variance} &  56.72 &	65.16 &	66.86 &	68.72 &	72.46 &	74.75 &	68.71 &	76.29    \\

\textbf{Percentil75} &  62.19 &	74.05 &	74.83 &	80.67 &	82.06 &	82.14 &	78.64 &	80.51  \\

\textbf{LBP} &    76.13 &	79.23 &	77.15 &	78.43 &	77.67 &	77.32 &	76.99 &	76.79  \\

      \textbf{Covariance} &   48.28 &	64.26 &	65.77 &	68.99 &	71.02 & 72.49 &	69.85 &	72.75   \\

      \textbf{GLCM} &   18.17 &	26.85 &	26.68 &	26.80 &	23.80 &	24.35 &	25.85 &	21.92    \\

      \textbf{Enhanced Fractal} &      77.19 &	81.48 &	80.11 &	83.21 &	83.46 &	82.94 &	\textbf{83.87} &	83.16 \\

\hline
\end{tabular}
\end{center}

\label{tab:Table3}
\end{table*}

\begin{table*}[!h]
\begin{center}\small
\caption{Results for Outext test suite 14 database.}
\begin{tabular}{ccccccccc}
\hline
                                                                        \multicolumn{ 9}{c}{ \textbf{Scales x Orientations} } \\
\hline
    &       &       &       &       &       &       &      &       \\
   \multicolumn{ 1}{c}{\textbf{Gabor +}} & \multicolumn{ 1}{c}{\textbf{2 x 6}} & \multicolumn{ 1}{c}{\textbf{3 x 4}} & \multicolumn{ 1}{c}{\textbf{3 x 5}} & \multicolumn{ 1}{c}{\textbf{4 x 4}} & \multicolumn{ 1}{c}{\textbf{4 x 6}} & \multicolumn{ 1}{c}{\textbf{5 x 5}} & \multicolumn{ 1}{c}{\textbf{6 x  3}} & \multicolumn{ 1}{c}{\textbf{6 x 6}} \\
\hline

    \textbf{Energy} &  27.94 &	44.82 &	44.73 &	48.39 &	49.73 &	50.06 &	48.38 &	52.20 \\

  \textbf{Variance} &   32.15 &	43.17 &	42.29 &	48.30 &	50.01 &	48.80 &	46.66 &	51.96  \\

\textbf{Percentil75} &   32.02 &	41.84 &	41.00 &	50.59 &	52.08 &	51.14 &	49.84 &	53.47 \\

\textbf{LBP} &    53.65 &	57.01 &	56.35 &	56.57 &	54.56 &	54.04 &	54.98 &	53.33  \\

      \textbf{Covariance} &   37.34 &	51.62 &	52.71 &	57.90 &	58.89 &	59.02 &	56.52 &	61.23   \\

      \textbf{GLCM} &   22.03 &	27.03 &	23.37 &	23.15 &	21.54 &	18.21 &	19.04 &	18.08   \\

      \textbf{Enhanced Fractal} &      63.28 &     \textbf{64.46} &      63.85 &      62.72 &      63.95 &     62.01 &      62.52 &      60.66 \\

\hline
\end{tabular}
\end{center}

\label{tab:Table4}
\end{table*}

\begin{table*}[!h]
\begin{center}\small
\caption{Results for Outext test suite 16 database.}
\begin{tabular}{ccccccccc}
\hline
                                                                        \multicolumn{ 9}{c}{ \textbf{Scales x Orientations} } \\
\hline
    &       &       &       &       &       &       &      &       \\
   \multicolumn{ 1}{c}{\textbf{Gabor +}} & \multicolumn{ 1}{c}{\textbf{2 x 6}} & \multicolumn{ 1}{c}{\textbf{3 x 4}} & \multicolumn{ 1}{c}{\textbf{3 x 5}} & \multicolumn{ 1}{c}{\textbf{4 x 4}} & \multicolumn{ 1}{c}{\textbf{4 x 6}} & \multicolumn{ 1}{c}{\textbf{5 x 5}} & \multicolumn{ 1}{c}{\textbf{6 x  3}} & \multicolumn{ 1}{c}{\textbf{6 x 6}} \\
\hline

    \textbf{Energy} &  36.32 &	61.53 &	60.20 &	66.20 &	66.60 &	66.29 &	64.63 &	66.38 \\

  \textbf{Variance} &   35.60 &	53.45 &	53.24 &	58.69 &	61.05 &	60.30 &	57.54 &	62.55 \\

\textbf{Percentil75} &  37.45 &	56.21 &	55.45 &	62.91 &	62.77 &	63.03 &	60.00 &	63.09 \\

\textbf{LBP} &   65.92 & 68.21 &	66.43 &	65.58 &	62.10 &	59.59 &	63.81 &	55.69   \\

      \textbf{Covariance} & 38.83 &	63.03 &	63.71 &	69.30 &	68.56 &	66.88 &	65.26 &	63.81 \\

      \textbf{GLCM} & 19.73 & 30.22 &	27.05 &	26.72 &	21.49 &	20.77 &	23.02 &	17.07  \\

      \textbf{Enhanced Fractal} &      69.58 &     74.12 &      73.90 &    \textbf{77.02} &   74.82    &     75.66  &     73.88 &      70.20 \\

\hline
\end{tabular}
\end{center}

\label{tab:Table5}
\end{table*}

\ref{tab:Table1} shows the results obtained for the full Brodatz database. The best result obtained by one of the compared methods (LBP) is $92.75\%$. The proposed method obtains $95.59\%$. Moreover, the results maintain a lower variability when more scales and orientations are used. In \ref{tab:Table2} The difference is much more significant. Our method obtains $91.58\%$ and the Gabor+LBP method obtains $86.17\%$. \ref{tab:Table3} shows the results for the Outex 5 classification test suit. The proposed method obtained $83.87\%$ and the Gabor+Percentil75 obtains $82.14\%$. The finest overall reported result for Outex 5 is $86\%$. \ref{tab:Table4} shows the results obtained for the Outex 14 classification suit. The proposed method obtains $64.46\%$ and the Gabor+Covariance method obtains $61.23\%$ where the best overall result reported for Outex 14 is $69\%$. Finally, \ref{tab:Table5} shows the results obtained for the Outex 16 classification suit. The proposed method obtains $77.02\%$ and the Gabor+Covariance method obtains $69.30\%$.

\section{Conclusions}

We have presented a novel technique that improves the Gabor wavelets to extract features from texture images. The effectiveness of the method is demonstrated by various experiments. The proposed method obtained the best results on all the image databases used. Texture feature extraction is a difficult task and it has been widely addressed but most of the approaches found in the literature only focus on a short range of texture conditions. The variability of the results of the compared methods shows the weakness of these methods when the image datasets used present a great intra-class variability a wide range of texture types and variations in the capture conditions. Different image datasets were selected with the purpose of presenting consistent results. However, this is not very common since most methods only perform well under tight image conditions. As shown in the results, most of the related methods only work well with one image dataset. Moreover, the variability of results on each compared method for a single dataset shows their sensibility to Gabor wavelets parameters. In this matter, the proposed method performs consistently in all experiments showing a clear independence of both methods and a successful conjunction to obtain rich texture features.

\section*{Acknowledgments}

A. Gomez Z. gratefully acknowledges the financial support of FAPESP (The State of Sao Paulo Research Foundation) Proc. 2009/04362.
J. B. Florindo gratefully acknowledges the financial support of FAPESP Proc. 2012/19143-3.
O. M. Bruno gratefully acknowledges the financial support of CNPq (National Council for Scientific and Technological Development, Brazil) (Grant \#308449/2010-0 and \#473893/2010-0) and FAPESP (The State of S\~ao Paulo Research Foundation) (Grant \# 2011/01523-1).

\section{References}


\end{document}